\documentclass[3p,authoryear]{elsarticle}

\usepackage{graphicx}
\graphicspath{{./}{figures/}}
\usepackage{multirow}
\usepackage{amsmath,amssymb,amsfonts}
\usepackage{xcolor}
\usepackage{textcomp}
\usepackage{booktabs}
\usepackage{placeins}
\usepackage{rotating}
\usepackage{xurl}
\usepackage[hidelinks]{hyperref}
\usepackage{orcidlink}

\journal{Waste Management}

\begin{document}

\begin{frontmatter}

\title{WasteBench: A Cross-Domain Benchmark for Waste Detection Across Conveyor, Litter, Aerial and Water Domains}

\author[ehu]{Salah Eddine Bekhouche\,\orcidlink{0000-0001-5538-7407}}

\author[cnr]{Fadi Abdeladhim Zidi\,\orcidlink{0009-0007-4791-8555}}

\author[biskra]{Ahmed Marouane Djouamaa\,\orcidlink{0009-0004-5947-884X}}

\author[cnr]{Abdellah Zakaria Sellam\,\orcidlink{0009-0003-6876-2220}}

\author[ehu]{Fadi Dornaika\,\orcidlink{0000-0001-6581-9680}\corref{cor1}}
\ead{fadi.dornaika@ehu.eus}
\cortext[cor1]{Corresponding author.}

\author[cnr]{Cosimo Distante\,\orcidlink{0000-0002-1073-2390}}

\author[umk]{Abdenour Hadid\,\orcidlink{0000-0001-9092-735X}}

\affiliation[ehu]{organization={University of the Basque Country (EHU)},
            addressline={Barrio Sarriena s/n},
            city={Leioa},
            postcode={48940},
            state={Bizkaia},
            country={Spain}}

\affiliation[cnr]{organization={Institute of Applied Sciences and Intelligent Systems, National Research Council of Italy (CNR)},
            addressline={Campus Ecotekne, Via Monteroni snc},
            postcode={73100},
            city={Lecce},
            country={Italy}}

\affiliation[biskra]{organization={Department of Computer Science, Mohamed Khider University},
            postcode={07000},
            city={Biskra},
            country={Algeria}}

\affiliation[umk]{organization={Faculty of Data Science and Computing, Universiti Malaysia Kelantan},
            postcode={16100},
            city={Kota Bharu},
            state={Kelantan},
            country={Malaysia}}

\begin{abstract}
Public datasets for waste detection are numerous but fragmented, and a detector is usually trained and tested on a single one of them. This hides how much accuracy a detector loses when it is moved to another sorting plant or to another type of scene, and what it costs to run. In this paper, we present WasteBench, a benchmark that unifies six public datasets from four domains, industrial conveyors, litter in the wild, aerial images and inland water, under a shared two-level taxonomy (waste versus background, and five material classes) with fixed splits. WasteBench defines five evaluation settings, from in-domain training to zero-shot and few-shot transfer, and applies them to three closed-set detectors, three open-vocabulary detectors and two vision-language models, together with their latency, memory and energy. The gap between datasets is severe: on average, a closed-set detector trained on one dataset keeps 18--48\% of its in-domain mean average precision (mAP@0.5) on the others, and conveyor datasets from different plants transfer only slightly better than different domains. Without target labels, the best open-vocabulary detector, OWLv2-Base, reaches a mean mAP@0.5 of 0.424, between YOLO26-s (0.368) and RF-DETR-small (0.453) trained on the other five datasets, at 24 times the GPU latency of YOLO26-s. Fine-tuning on 100 labeled images of the target dataset recovers 77--93\% of the in-domain accuracy. A detector moved to a new facility or monitoring site should therefore be planned together with a small local labeling effort. The taxonomy, splits and loaders are available at \url{github.com/Bekhouche/WasteBench}.
\end{abstract}


\end{frontmatter}

\section{Introduction}\label{sec:intro}

Automatic waste detection locates and classifies discarded items in an image. It is used at several points of the waste management chain: on the conveyors of sorting plants, where a detection can trigger the separation of an item, in the monitoring of litter in streets and public spaces, in drone surveys and in the surveillance of floating debris on rivers and lakes. Depending on the downstream task, it is formulated as a binary problem, waste against background, or as a material classification problem, plastic against metal against glass.

Public datasets for waste detection have multiplied in the last few years, covering industrial conveyors (SortWaste, ZeroWaste-f, WaRP-D), litter photographed in the wild (TACO), aerial views from drones (UAVVaste) and floating debris on inland water (IWHR\_AI\_Lable\_Floater\_V1, IWHR-Floater in the following). Each one was collected with its own camera setup, label set and, in most cases, evaluation protocol, and a detector is almost always trained and tested on a single one of them. For a plant operator or a municipality, this hides two practical problems. The first is the gap between datasets: a detector trained on the conveyor of one plant is rarely evaluated on another plant, on street litter or on drone imagery, although a system that is reused at a new site faces exactly this change. The second is the deployment cost: latency, memory and power are reported for some single-dataset systems \citep{uavvaste2021,cordova2022plastopol}, but not alongside cross-dataset accuracy, so it remains unclear which models can run on the devices of a sorting line, a drone or a monitoring station. Open-vocabulary detectors and vision-language models (VLMs) promise detection without task-specific training, but there is no systematic evidence of how well they handle the clutter, deformable shapes, translucency and small sizes of waste, nor of what they cost to run.

In this paper, we present WasteBench, a benchmark built to answer four research questions. RQ1 asks how much accuracy a detector loses when it is trained on one dataset and tested on another, within the conveyor domain and across domains. RQ2 asks whether open-vocabulary detectors and VLMs close this gap without any label of the target dataset, and how many labeled target images a detector needs, depending on its starting point. RQ3 asks what deployment cost each model family adds for its accuracy, and RQ4 how object size and scene density affect missed detections.

The main contributions of this work are as follows. First, we unify six public waste datasets under a shared two-level taxonomy, waste versus background and five material classes, with every native label mapped and justified in a table released with the benchmark. Second, a five-setting evaluation protocol, from in-domain training to zero-shot and few-shot transfer, is applied to eight models from three families: closed-set detectors, open-vocabulary detectors and VLMs used as grounding detectors. Third, deployment cost (latency, peak memory, GPU power and energy per image) is measured under the same protocol as accuracy. Finally, we release the taxonomy tables, the split files and the loaders, together with scripts that download each dataset from its original source and convert it to the shared taxonomy and splits.

The rest of the paper is organized as follows. Section~\ref{sec:related} reviews waste datasets and the detector families we evaluate. Section~\ref{sec:wastebench} describes the datasets, the taxonomy and the protocol, and Section~\ref{sec:models} the experimental setup. Sections~\ref{sec:results} to \ref{sec:failure} report the accuracy, cost and failure results, Section~\ref{sec:discussion} discusses their implications for waste management practice and their limitations, and Section~\ref{sec:conclusion} concludes. Supplementary tables, figures and notes are numbered with the prefix S.

\section{Related work}\label{sec:related}

\subsection{Waste detection datasets}\label{sec:related-datasets}

Three public datasets were collected on the conveyors of different sorting plants. SortWaste \citep{sortwaste2026} densely annotates 5,261 images with 87,252 objects across eight classes, and its authors report that a plastic-only subtask (mAP of 0.597) remains far from what cluttered, overlapping scenes require. ZeroWaste-f \citep{zerowaste2022} provides instance masks over four material classes and frames the task as the segmentation of deformable, often translucent objects in extreme clutter. WaRP-D \citep{warp2023} has the finest native taxonomy of the three, 28 classes grouped into five superclasses, and its authors identify class imbalance as a limiting factor. Outside the plant, TACO \citep{taco2020} is a crowdsourced dataset of litter in open scenes with 60 fine classes and no official split, PlastOPol \citep{cordova2022plastopol} adds 2,418 litter images and reports the processing time and memory of several detectors on a smartphone, and UAVVaste \citep{uavvaste2021} collects 772 low-altitude drone images of a single class, rubbish, with speed measurements on embedded devices. For inland water, IWHR-Floater \citep{iwhrfloater2025} provides 3,000 images of floating debris from shore-based cameras and handheld devices, and FloW \citep{cheng2021flow} is a floating-waste benchmark captured from an unmanned surface vehicle, on which \citet{sumon2026floating} compared YOLO, RT-DETR and Faster R-CNN with four ensembling strategies.

Each of these papers evaluates its dataset in isolation, and none reports a cross-dataset transfer matrix. The work closest to ours is detect-waste \citep{detectwaste2022}, which pools seven litter datasets, including TACO and UAVVaste, into a single training set and reaches an mAP@0.5 of 0.655 in a binary litter-versus-background setting; its preprint version \citep{detectwastearxiv2021} also reports the result on each source dataset separately. Both evaluations remain in-distribution: no dataset is held out, and neither transfer nor deployment cost is reported. The VisDA 2022 challenge \citep{bashkirova2023visda} studies unsupervised domain adaptation for waste segmentation on conveyor images, with an unlabeled target set available during training. WasteBench differs in three ways: it measures detection across four domains, it withholds the target dataset entirely, and it adds open-vocabulary detectors, VLMs and deployment cost. Other waste work, from container recognition \citep{ramirez2020dumpsters} to recyclable-waste classification \citep{sayem2025enhancing,dahy2025recognizing}, follows the same single-dataset pattern, and a recent survey \citep{sur2026wastesurvey} names fragmented benchmarks and inconsistent protocols as open problems. Illegal dumping has also been studied as a video event from fixed cameras \citep{greco2026illegal,bouwmans2026illegal,pascarella2023aisustainability}, which is complementary to detecting the waste present in a single image.

\subsection{Detectors, open-vocabulary detectors and vision-language models}\label{sec:related-closedset}\label{sec:related-openvocab}\label{sec:related-vlm}

Faster R-CNN \citep{fasterrcnn2015} remains a standard two-stage baseline. YOLO26 \citep{yolo26_2026} is the current entry in the YOLO family, with an end-to-end head aimed at edge deployment, and RF-DETR \citep{rfdetr2026} is a real-time detection transformer built on a DINOv2 backbone, whose authors report strong results on RF100-VL, a cross-domain benchmark of 100 datasets. Such detectors have also been studied for robustness and cost in other cluttered scenes, such as autonomous driving \citep{kezzal2025deformable,benderradji2025yolo11} and car damage detection \citep{sellam2025cdiffdet}. Open-vocabulary detectors take the class names as text prompts at inference: Grounding DINO \citep{groundingdino2024} combines the DINO detector with grounded pre-training, OWLv2 \citep{owlv2_2023} scales this approach with self-training on image-text pairs, and YOLOE \citep{yoloe2025} brings text, visual and prompt-free modes to the YOLO family. For low-level nuclear waste, \citet{rojas2024nuclear} report that a closed-set detector is more accurate while the open-vocabulary OWL-ViT adapts more easily to new waste types. Florence-2 \citep{florence2_2024} and Qwen3-VL \citep{qwen3vl2025} are VLMs that can output bounding boxes from a text query; VLMs have also been applied to other fine-grained recognition tasks, such as pedestrian attribute recognition \citep{sellam2025vlmpar}.

\subsection{Domain generalization and efficient detection}\label{sec:related-domaingen}\label{sec:related-efficient}

Unsupervised domain adaptation for detection \citep{oza2021uda} assumes access to unlabeled target images during training, and recent surveys of cross-domain detection \citep{cdod2026} report that domain gaps remain severe even within a single application. Our leave-one-out setting is closer to domain generalization, since the target dataset is withheld entirely. Moreover, the usual benchmarks of this literature change the capture condition while keeping the object categories, whereas waste datasets change both at once. For deployment, lightweight detection relies on pruning, quantization, distillation and compact backbones \citep{mittal2024edge}, but it is rarely evaluated on small, cluttered or deformable objects.

\section{WasteBench: datasets, taxonomy and protocol}\label{sec:wastebench}

\subsection{Datasets}\label{sec:wastebench-datasets}

\begin{figure}[h]
\centering
\includegraphics[width=\textwidth]{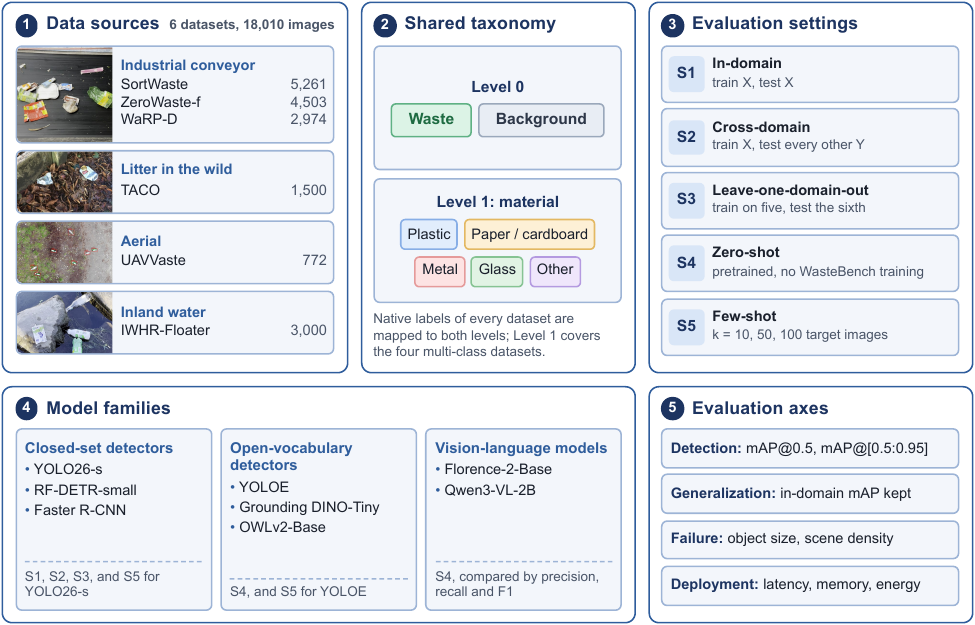}
\caption{WasteBench overview: six public datasets unified under one two-level taxonomy, evaluated under a five-setting protocol (S1--S5) across three model families, with detection, generalization, failure and deployment measured as separate axes.}\label{fig:overview}
\end{figure}

WasteBench unifies six public datasets from four domains (Table~\ref{tab:datasets}): three industrial conveyor datasets collected in three different plants, SortWaste \citep{sortwaste2026}, ZeroWaste-f \citep{zerowaste2022} and WaRP-D \citep{warp2023}; TACO \citep{taco2020} for litter in the wild; UAVVaste \citep{uavvaste2021} for aerial images; and IWHR-Floater \citep{iwhrfloater2025} for inland water. Since litter, aerial and water are each represented by a single dataset, a result for one of these domains is a result for that dataset. Figure~\ref{fig:overview} gives an overview of the benchmark.

We included public datasets that annotate individual waste objects with boxes or masks, with at least one dataset per domain. StreetView-Waste \citep{paulo2025streetviewwaste} is distributed under a data-license agreement and labels municipal bin types rather than loose waste, FloW \citep{cheng2021flow} is distributed through an application process, MJU-Waste \citep{wang2020mjuwaste} shows hand-held objects photographed indoors, TrashCan \citep{hong2020trashcan} covers underwater debris, a domain outside the four studied here, and AerialWaste \citep{torres2023aerialwaste} labels landfill sites rather than individual objects. PlastOPol \citep{cordova2022plastopol} and TrashCan are natural extensions of the benchmark.

\FloatBarrier

\begin{sidewaystable}[h]
\centering
\caption{The six datasets unified in WasteBench, with the splits used in our protocol. Object counts are those of the processed WasteBench release and differ slightly from the original papers for SortWaste (87,525 against 87,252) and UAVVaste (3,718 against 3,716). $^{\dagger}$Official train and test splits; WaRP-D has no official validation split, so ours is held out from its training split (Section~\ref{sec:wastebench-datasets}).}\label{tab:datasets}
\begin{tabular}{@{}llccccl@{}}
\toprule
Domain & Dataset & Classes & Images (tr/val/test) & Objects (tr/val/test) & Split & License \\
\midrule
Industrial conveyor & SortWaste    & 8  & 3{,}705 / 780 / 776   & 61{,}842 / 13{,}065 / 12{,}618 & Official  & CC BY 4.0 \\
Industrial conveyor & ZeroWaste-f  & 4  & 3{,}002 / 572 / 929   & 18{,}002 / 3{,}687 / 5{,}077   & Official  & CC BY-NC 4.0 \\
Industrial conveyor & WaRP-D       & 28 & 2{,}215 / 237 / 522   & 8{,}021 / 802 / 1{,}551        & Official$^{\dagger}$ & CC0 \\
Litter in the wild  & TACO         & 60 & 1{,}099 / 201 / 200   & 3{,}319 / 814 / 651            & Generated & CC BY 4.0 \\
Aerial              & UAVVaste     & 1  & 590 / 94 / 88         & 2{,}839 / 539 / 340            & Generated & Apache 2.0 \\
Inland water        & IWHR-Floater & 1 & 2{,}100 / 450 / 450 & 16{,}432 / 3{,}674 / 3{,}586 & Generated & Apache 2.0 \\
\bottomrule
\end{tabular}
\end{sidewaystable}
\FloatBarrier

TACO, UAVVaste and IWHR-Floater do not define an official split, so we generate fixed, seeded splits with a 70/15/15 target. For TACO and UAVVaste, images are first grouped by capture sequence, so that near-duplicate frames of one sequence never appear on both sides of a split; IWHR-Floater provides no sequence information and is split at the image level. WaRP-D has no official validation split, so we hold out 237 of its 2,452 training images for validation. A perceptual-hash check over all 18,010 images finds no image shared between two datasets. Within datasets, 49 groups of identical images cross a split boundary, 37 of them in IWHR-Floater; removing the affected IWHR-Floater test images does not lower its score (Supplementary Note~\ref{note:dedup}). Frames of the same fixed camera that are not identical can still fall on both sides of the IWHR-Floater split, so its in-domain results may be optimistic.

\subsection{Taxonomy}\label{sec:taxonomy}

WasteBench maps every dataset to two label levels. Level 0 is binary, waste against background, and every dataset participates at this level without any mapping decision. Level 1 has five material classes, plastic, paper/cardboard, metal, glass and other, plus an ignore label for boxes with no reliable material signal. These shared classes replace the native classes of each dataset, from 1 to 60 (Table~\ref{tab:datasets}), so the 60 TACO categories, the 28 WaRP-D classes and the 8 SortWaste classes are each mapped to one material, and every model, including the zero-shot ones, is evaluated on the mapped labels. UAVVaste and IWHR-Floater map entirely to ignore at Level 1, since their single classes carry no material information, and are evaluated at Level 0 only. We release one mapping table per dataset, with a justification for every non-obvious decision; for instance, WaRP groups glass bottles under its bottle superclass, which we map by material, to glass. Datasets with only a subset of materials, such as ZeroWaste-f, which has no glass class, are evaluated on the classes they contain.

Level 0 removes the material vocabulary, but not every difference in labeling. The datasets differ in annotation scope: in ZeroWaste-f and WaRP-D, similar material in the background can remain unlabeled, whereas SortWaste and TACO aim to annotate every waste object in the image. A detector trained on one scope is penalized on another for boxes that are correct but unlabeled, which contributes to the cross-dataset gap at both levels.

\subsection{Evaluation protocol}\label{sec:protocol}

WasteBench defines five settings (Table~\ref{tab:settings}), and every reported result is computed on a test split. S1 (in-domain) trains on the training split of a dataset X, selects the checkpoint on its validation split and tests on its test split. S2 (pairwise cross-domain) tests each S1 checkpoint, without retraining, on the test split of every other dataset Y of the same level. S3 (leave-one-domain-out) trains on the union of the training splits of all other datasets of the level and tests on Y, of which no split is used before testing. S4 (zero-shot) uses no WasteBench data for training or selection. S5 (few-shot) fine-tunes on $k=10$, 50 or 100 images of the training split of Y, starting from the S3 YOLO26-s, the COCO YOLO26-s or YOLOE, and selects the checkpoint on the full validation split of Y.

S2 contains two kinds of pairs: six pairs between two conveyor datasets, which change the plant but not the domain, and the remaining pairs, which also change the domain. We report the two kinds separately. Likewise, S3 removes one dataset: for TACO, UAVVaste and IWHR-Floater, this removes the domain entirely, whereas for a conveyor dataset the other two conveyor datasets remain in the training pool, so these rows measure transfer to a new plant.

\section{Experimental setup}\label{sec:models}

\subsection{Models}\label{sec:model-families}

WasteBench evaluates eight models from three families (Table~\ref{tab:models}). A closed-set detector predicts only the classes fixed at training and must be trained on labeled boxes of these classes. An open-vocabulary detector receives its classes at inference as text prompts, so it can detect a class without any labeled example from WasteBench. A VLM answers a text instruction about an image; it is not built for detection, but it can write box coordinates in its answer.

YOLO26-s \citep{yolo26_2026} is the primary closed-set detector, evaluated under S1, S2, S3 and S5, where it is fine-tuned either from its S3 checkpoint or from its COCO weights, that is, the weights released after pretraining on COCO \citep{lin2014coco}, a general-purpose detection dataset of 80 everyday object categories. RF-DETR-small \citep{rfdetr2026} and Faster R-CNN \citep{fasterrcnn2015} with a ResNet-50-FPN v2 backbone are evaluated under S1, S2 and S3 (Faster R-CNN at Level 0 only under S3). Level 0 is evaluated on all six datasets and Level 1 on the four datasets with material labels. The open-vocabulary detectors Grounding DINO-Tiny \citep{groundingdino2024}, OWLv2-Base \citep{owlv2_2023} and YOLOE \citep{yoloe2025}, built on YOLO26-s and used with text prompts, and the VLMs Florence-2-Base \citep{florence2_2024} and Qwen3-VL-2B \citep{qwen3vl2025} are evaluated under S4 without any fine-tuning; YOLOE is also fine-tuned under S5. The open-vocabulary detectors receive several synonyms per class in one forward pass and are evaluated on the full test splits. The VLMs receive one phrase per class, one generation per phrase, and are evaluated on a fixed sample of up to 100 test images per dataset, since they are slower. The prompts are fixed for each model and level (Tables~\ref{tab:prompts} and~\ref{tab:prompt-inputs}); they were informed by examining evaluation images, although no model weights were fitted to WasteBench data (Supplementary Note~\ref{note:prompts}).

\subsection{Metrics}\label{sec:metrics}

We report mAP at an IoU threshold of 0.5 (mAP@0.5) and averaged over thresholds from 0.5 to 0.95 (mAP@[0.5:0.95]), separately for Level 0 and Level 1. For the YOLO26-s results, we estimate 95\% confidence intervals from 1,000 bootstrap resamples of the test images, and we describe a difference with another model as clear only when that model's score lies outside the YOLO26-s interval. The VLMs return boxes without a confidence score, so their boxes cannot be ranked and AP, which measures the ranking of the boxes, is not defined for them. We therefore do not report mAP for the VLMs. Their output is a single operating point, and we evaluate it by precision, recall and F1 at IoU 0.5, with the open-vocabulary detectors read at a confidence of 0.25 on the same images (Supplementary Note~\ref{note:vlm-ap}).

For S2, the fraction of in-domain accuracy kept is the mean cross-dataset mAP@0.5 divided by the mean in-domain mAP@0.5 over the datasets of a level; for S3, it is the held-out mAP@0.5 divided by the S1 result of the same detector on that dataset. For the failure analysis, we report recall at IoU 0.5 for predictions above a confidence of 0.25, with ground-truth boxes split into terciles by relative box area and by the number of annotated objects in their image, both computed within each dataset. Deployment cost is measured at batch size 1 for the three closed-set detectors and OWLv2-Base, on an Intel Xeon Silver 4314 server and a single NVIDIA L4 GPU, as the median latency of 50 runs, the throughput, the peak memory, the GPU board power and the GPU energy per image (Supplementary Note~\ref{note:deployment}).

\subsection{Implementation}\label{sec:implementation}

Each closed-set detector is trained with its official implementation: Ultralytics for YOLO26-s, the released RF-DETR code and torchvision for Faster R-CNN. YOLO26-s and RF-DETR-small use up to 100 epochs at an input size of 640 with an early-stopping patience of 30, whereas Faster R-CNN uses up to 30 epochs with a patience of 10 and a shorter side of 800 pixels; batch sizes and optimizers follow the defaults of each implementation. The comparisons between architectures are therefore comparisons of complete training recipes. Each configuration is trained once with a fixed seed, and the YOLO26-s S1 training is repeated with two further seeds to estimate the variation between runs. All configurations, prompts and hyperparameters are released with the code.

\section{Results}\label{sec:results}

\subsection{In-domain and leave-one-domain-out accuracy}\label{sec:s1-results}

\begin{table}[h]
\centering
\caption{In-domain (S1) and leave-one-domain-out (S3) results of the three closed-set detectors, mAP@0.5, test split. Kept is the S3 result as a fraction of the same detector's S1 result on that dataset. Faster R-CNN is evaluated under S3 at Level 0 only. Bold marks the best detector per row, separately for S1 and S3. The 95\% bootstrap intervals and the three-seed results of YOLO26-s are given in Table~\ref{tab:yolo-uncertainty}.}\label{tab:s1s3}
\footnotesize
\setlength{\tabcolsep}{3pt}
\begin{tabular}{@{}llccccccccc@{}}
\toprule
 & & \multicolumn{3}{c}{YOLO26-s} & \multicolumn{3}{c}{Faster R-CNN} & \multicolumn{3}{c}{RF-DETR-small} \\
Dataset & Level & S1 & S3 & Kept & S1 & S3 & Kept & S1 & S3 & Kept \\
\midrule
SortWaste    & 1 & 0.530 & 0.304 & 57\% & 0.556 & -- & -- & \textbf{0.720} & \textbf{0.441} & 61\% \\
ZeroWaste-f  & 1 & 0.404 & 0.094 & 23\% & 0.499 & -- & -- & \textbf{0.570} & \textbf{0.243} & 43\% \\
WaRP-D       & 1 & 0.564 & 0.086 & 15\% & 0.551 & -- & -- & \textbf{0.724} & \textbf{0.142} & 20\% \\
TACO         & 1 & 0.164 & 0.026 & 16\% & 0.190 & -- & -- & \textbf{0.348} & \textbf{0.098} & 28\% \\
SortWaste    & 0 & 0.858 & 0.696 & 81\% & 0.894 & 0.675 & 76\% & \textbf{0.917} & \textbf{0.791} & 86\% \\
ZeroWaste-f  & 0 & 0.567 & 0.138 & 24\% & 0.627 & 0.132 & 21\% & \textbf{0.684} & \textbf{0.217} & 32\% \\
WaRP-D       & 0 & 0.732 & 0.105 & 14\% & 0.756 & \textbf{0.342} & 45\% & \textbf{0.815} & 0.217 & 27\% \\
TACO         & 0 & 0.439 & 0.219 & 50\% & 0.475 & 0.299 & 63\% & \textbf{0.490} & \textbf{0.390} & 80\% \\
UAVVaste     & 0 & 0.640 & 0.577 & 90\% & 0.763 & \textbf{0.608} & 80\% & \textbf{0.807} & 0.480 & 59\% \\
IWHR-Floater & 0 & \textbf{0.914} & 0.472 & 52\% & 0.894 & 0.579 & 65\% & \textbf{0.914} & \textbf{0.625} & 68\% \\
\bottomrule
\end{tabular}
\end{table}
\FloatBarrier

Table~\ref{tab:s1s3} reports the in-domain (S1) and leave-one-domain-out (S3) results of the three closed-set detectors. In-domain, RF-DETR-small is the strongest detector on nine of the ten configurations, by up to 0.19 mAP@0.5 over YOLO26-s (SortWaste, Level 1), and ties with it on IWHR-Floater (0.914). Its advantage is clear on eight of these nine configurations, TACO at Level 0 being within the YOLO26-s interval. Faster R-CNN is above YOLO26-s on eight configurations, clearly on five. Over the three seeds, the YOLO26-s results differ by at most 0.027 on nine configurations and by 0.068 on ZeroWaste-f at Level 1, and every difference that we describe as clear holds against all three seeds (Table~\ref{tab:yolo-uncertainty}). Every dataset is easier at Level 0, which does not require naming the material. TACO is the weakest dataset at both levels, so its difficulty is not only an artifact of the taxonomy, and metal, the smallest Level 1 class, is detected poorly on SortWaste (recall of 0.123 for 215 instances; Table~\ref{tab:s1-sortwaste}).

Pooling five source datasets under S3 closes only part of the gap. For YOLO26-s, the mean S3 mAP@0.5 is 0.128 at Level 1 and 0.368 at Level 0, and it is 0.439 for Faster R-CNN and 0.453 for RF-DETR-small at Level 0. SortWaste and UAVVaste keep the most when held out (81\% and 90\% for YOLO26-s), whereas ZeroWaste-f and WaRP-D keep only 14\% to 24\%, although two other conveyor datasets remain in their training pool. The worst held-out dataset at Level 0 reaches 0.105 for YOLO26-s (WaRP-D), 0.132 for Faster R-CNN and 0.217 for RF-DETR-small.

\subsection{Pairwise transfer}\label{sec:s2-s5-results}

\begin{figure}[h]
\centering
\includegraphics[width=\textwidth]{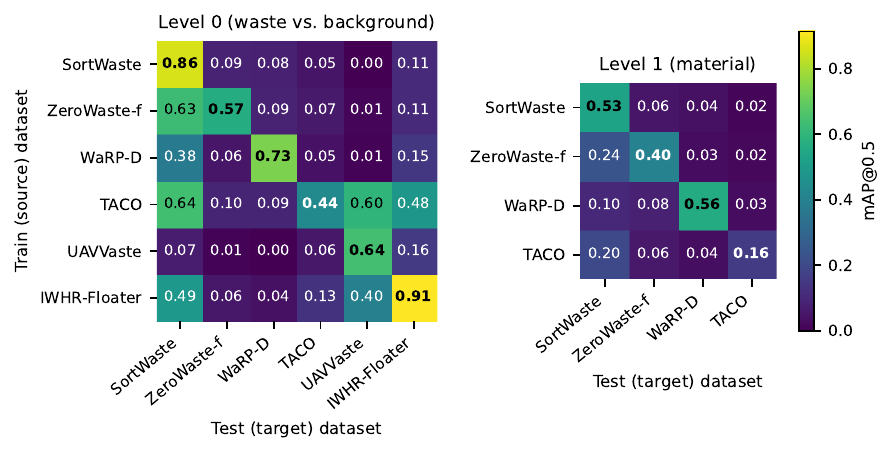}
\caption{S1 in-domain accuracy (diagonal, bold) and S2 pairwise transfer (off-diagonal), YOLO26-s, mAP@0.5. Every row is the checkpoint trained on one dataset, tested on every test split; SortWaste, ZeroWaste-f and WaRP-D form the conveyor domain.}\label{fig:s2-heatmap}
\end{figure}
\FloatBarrier

Figure~\ref{fig:s2-heatmap} shows the complete S2 matrices of YOLO26-s. The gap between datasets is severe at both levels: the mean S2 mAP@0.5 is 0.076 at Level 1 and 0.173 at Level 0, against in-domain means of 0.416 and 0.692, so a detector applied to another dataset keeps only 18\% of its in-domain accuracy at Level 1 and 25\% at Level 0 (95\% intervals of 17.4--19.8\% and 24.5--25.9\%; 17.7 $\pm$ 1.8\% and 25.4 $\pm$ 0.3\% over three seeds). The pairs range from a near-total failure, UAVVaste to WaRP-D at Level 0 (0.002), to TACO to SortWaste (0.636). On average, no detector keeps more than half of its in-domain accuracy, although single pairs can exceed this level (TACO to SortWaste, 0.636 at Level 0 for YOLO26-s): at Level 0, Faster R-CNN keeps 38\% and RF-DETR-small 48\% (Table~\ref{tab:s2-models}), and the mean single-source transfer of RF-DETR-small (0.370) is close to what YOLO26-s reaches from five pooled datasets (0.368). We can observe that the gain from pooling decreases as single-source transfer improves: at Level 0, S3 raises the mean S2 result by a factor of 2.1 for YOLO26-s, 1.6 for Faster R-CNN and 1.2 for RF-DETR-small.

Transfer within the conveyor domain is better than across domains for all three detectors, and the difference is clear for YOLO26-s at both levels, but it is small compared with the loss itself (Table~\ref{tab:s2-within}). A detector moved to another plant keeps only 31\%, 40\% and 47\% of its conveyor in-domain accuracy at Level 0 (YOLO26-s, Faster R-CNN, RF-DETR-small), and four of the six Level 0 conveyor pairs of YOLO26-s stay below 0.10 mAP@0.5. Most of the gap therefore separates datasets, even from the same domain. Two dataset-specific causes add to the change of camera and background: at Level 1, not every dataset contains every material (SortWaste has no glass), and at both levels, correct detections of unlabeled objects count as false positives on ZeroWaste-f and WaRP-D. Our protocol does not separate these causes. The patterns of the individual source and target datasets are given in Supplementary Note~\ref{note:s2-patterns}.

\subsection{Zero-shot detection}\label{sec:s4-results}

\begin{table}[h]
\centering
\caption{S4 zero-shot, mAP@0.5 per dataset, test split. YOLOE (labels) uses the literal taxonomy labels as prompts; every other zero-shot row uses the concrete-noun prompts of Table~\ref{tab:prompts}. The two VLMs give no confidence score and therefore no mAP; they are compared with the open-vocabulary detectors by precision, recall and F1 in Supplementary Table~\ref{tab:s4-matched} (Section~\ref{sec:metrics}). Bold marks the best open-vocabulary detector per column. The two YOLO26-s rows are references from Table~\ref{tab:s1s3}, not zero-shot results: S1 is trained on the target dataset, S3 on the five other datasets.}\label{tab:s4-summary}
\footnotesize
\setlength{\tabcolsep}{3.5pt}
\begin{tabular}{@{}lccccccc@{}}
\toprule
Model & SortWaste & ZeroWaste-f & WaRP-D & TACO & UAVVaste & IWHR-Floater & Mean \\
\midrule
\multicolumn{8}{@{}l}{\textit{Level 0}} \\
YOLOE (labels)          & 0.000 & 0.010 & 0.000 & 0.010 & 0.000 & 0.010 & 0.005 \\
YOLOE                   & 0.049 & 0.030 & 0.071 & 0.092 & 0.030 & 0.148 & 0.070 \\
Grounding DINO-Tiny     & 0.401 & 0.139 & 0.234 & 0.241 & 0.364 & 0.413 & 0.299 \\
OWLv2-Base              & \textbf{0.478} & \textbf{0.179} & \textbf{0.327} & \textbf{0.316} & \textbf{0.685} & \textbf{0.560} & \textbf{0.424} \\
\textit{YOLO26-s, S1}   & \textit{0.858} & \textit{0.567} & \textit{0.732} & \textit{0.439} & \textit{0.640} & \textit{0.914} & \textit{0.692} \\
\textit{YOLO26-s, S3}   & \textit{0.696} & \textit{0.138} & \textit{0.105} & \textit{0.219} & \textit{0.577} & \textit{0.472} & \textit{0.368} \\
\midrule
\multicolumn{8}{@{}l}{\textit{Level 1}} \\
YOLOE (labels)          & 0.021 & 0.017 & 0.010 & 0.027 & -- & -- & 0.019 \\
YOLOE                   & 0.045 & 0.023 & 0.028 & 0.099 & -- & -- & 0.049 \\
Grounding DINO-Tiny     & 0.197 & 0.091 & 0.098 & 0.132 & -- & -- & 0.130 \\
OWLv2-Base              & \textbf{0.259} & \textbf{0.141} & \textbf{0.109} & \textbf{0.262} & -- & -- & \textbf{0.193} \\
\textit{YOLO26-s, S1}   & \textit{0.530} & \textit{0.404} & \textit{0.564} & \textit{0.164} & -- & -- & \textit{0.416} \\
\textit{YOLO26-s, S3}   & \textit{0.304} & \textit{0.094} & \textit{0.086} & \textit{0.026} & -- & -- & \textit{0.128} \\
\bottomrule
\end{tabular}
\end{table}
\FloatBarrier

Prompt wording matters strongly for YOLOE: with the literal taxonomy labels, its mean mAP@0.5 is 0.005 at Level 0, and concrete-noun prompts such as ``trash'', ``plastic bottle'' and ``cardboard box'' raise it to 0.070 (Table~\ref{tab:s4-summary}). OWLv2-Base is the strongest open-vocabulary detector on every dataset at both levels, with means of 0.424 at Level 0 and 0.193 at Level 1, followed by Grounding DINO-Tiny (0.299 and 0.130). Its transfer is uneven: 0.179 on ZeroWaste-f, but 0.685 on UAVVaste, comparable to the in-domain YOLO26-s (0.640). This may be due to the sparse scenes of UAVVaste (4.8 objects per image against 16.6 for SortWaste) and to its single class, rubbish, which matches generic prompts such as ``trash'', whereas on a conveyor every object is waste by context only; we did not test these explanations directly.

The S3 results give the fairer reference, since a model trained on the five other datasets has not seen the target dataset either. OWLv2-Base is ahead of the S3 YOLO26-s on five of the six datasets at Level 0 (mean 0.424 against 0.368), but the S3 RF-DETR-small (0.453) is ahead of OWLv2-Base on four of them, and the S3 Faster R-CNN reaches a similar mean (0.439). The two VLMs are compared with the open-vocabulary detectors on the same 100-image samples by F1 at IoU 0.5 (Supplementary Table~\ref{tab:s4-matched}). At Level 0, both remain behind OWLv2-Base (F1 of 0.452) and Grounding DINO-Tiny (0.313). Florence-2-Base returns at least one box per query, even for an absent object, whereas Qwen3-VL-2B returns an empty list when the object is absent, which gives it a higher precision (0.393 against 0.337) and F1 (0.249 against 0.172). At Level 1, Qwen3-VL-2B reaches an F1 of 0.326, above Grounding DINO-Tiny (0.241) and close to OWLv2-Base (0.363).

\subsection{Few-shot adaptation}\label{sec:s5-results}

S5 measures how many labeled target images a detector needs, depending on where it starts: YOLO26-s from its S3 checkpoint, YOLO26-s from its COCO weights, and YOLOE, whose head starts from the text embeddings of the class names. The subsets of $k$ images are nested, and every run selects its checkpoint and its learning rate on the full target validation split, so the labeled target data seen by a run is $k$ training images plus 94 to 780 validation images (Supplementary Note~\ref{note:fewshot}).

\begin{figure}[h]
\centering
\includegraphics[width=\textwidth]{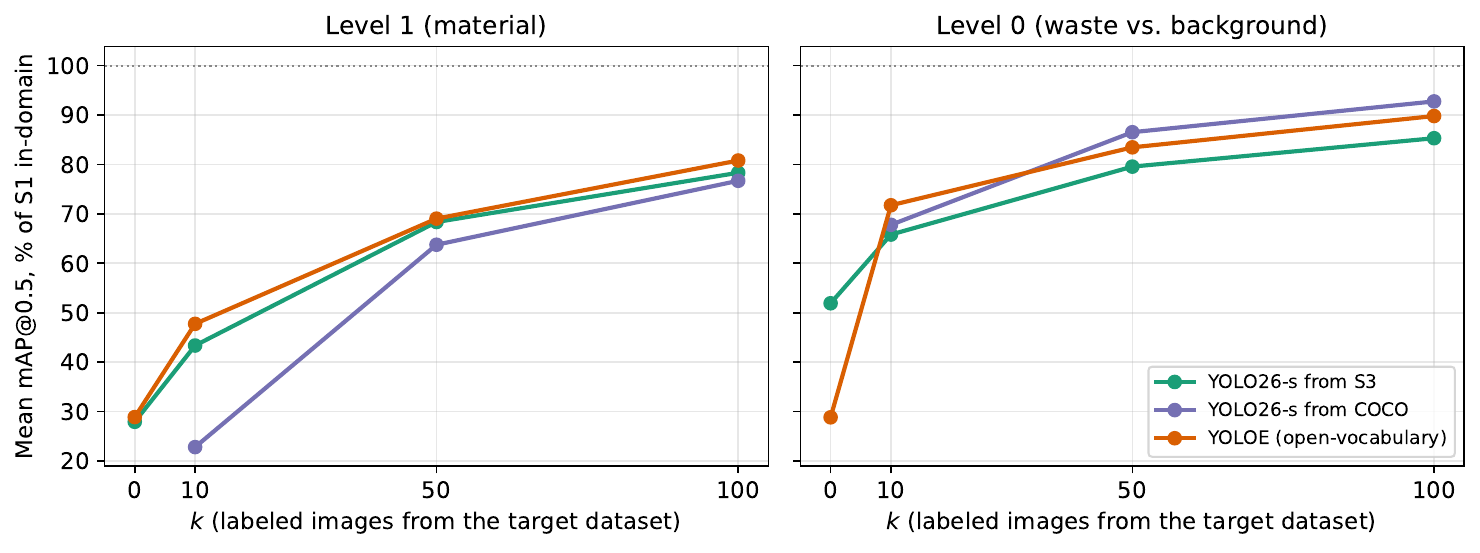}
\caption{S5 few-shot recovery by starting point: mean over the datasets of each level of each result (mAP@0.5) as a percentage of its YOLO26-s S1 in-domain result (a mean of per-dataset ratios), against the number $k$ of labeled target images. The S3 and YOLOE curves start from their $k=0$ results, the S3 result of YOLO26-s and the zero-shot result of YOLOE; the COCO start has no $k=0$ result.}\label{fig:s5-curves}
\end{figure}
\FloatBarrier

We can observe in Figure~\ref{fig:s5-curves} that a hundred target training images recover most of the in-domain accuracy from any starting point: 85\% from S3, 93\% from COCO and 90\% from YOLOE at Level 0, and 78\%, 77\% and 81\% at Level 1 (Table~\ref{tab:s5-results}). The starting point matters mainly at the smallest budgets. At Level 1 and $k=10$, the S3 start recovers 43\% of the in-domain accuracy against 23\% for the COCO start, but at $k=100$ the two recover almost the same fraction (78\% and 77\%), so the benefit of waste-specific pretraining for the material classes is limited to small budgets. At Level 0, the S3 start brings no advantage beyond $k=10$, and on average YOLOE is the strongest start at $k=10$ at both levels (72\% and 48\%). A few runs exceed the S1 reference, such as SortWaste at Level 1 from $k=50$ with the S3 start (0.573 against 0.530, outside the S1 interval), which suggests that the single-run S1 baseline, trained without a learning-rate search, is not an upper bound.

\section{Deployment cost}\label{sec:efficiency}

\begin{table}[h]
\centering
\caption{Latency, throughput, peak memory, GPU board power and GPU energy per image at batch size 1 (median of 50 runs). CPU: Intel Xeon Silver 4314. GPU: single NVIDIA L4. Energy is the mean board power times the median latency, idle power included; CPU power is not measured. INT8 is reported for YOLO26-s only.}\label{tab:efficiency}
\footnotesize
\setlength{\tabcolsep}{4pt}
\begin{tabular}{@{}llcccccc@{}}
\toprule
Model & Device & Precision & Latency (ms) & Thr. (fps) & Mem (MB) & Power (W) & Energy (J) \\
\midrule
YOLO26-s            & CPU & FP32 & 44.7   & 22.4 & 781  & -- & -- \\
YOLO26-s            & GPU & FP32 & 11.5   & 86.8 & 98   & 28.4 & 0.33 \\
YOLO26-s            & GPU & FP16 & 17.4   & 57.6 & 105  & 31.1 & 0.54 \\
YOLO26-s            & CPU & INT8 & 153.7  & 6.5  & 876  & -- & -- \\
RF-DETR-small       & CPU & FP32 & 133.5  & 7.5  & 1187 & -- & -- \\
RF-DETR-small       & GPU & FP32 & 17.1   & 58.7 & 331  & 35.3 & 0.60 \\
RF-DETR-small       & GPU & FP16 & 21.1   & 47.5 & 380  & 34.1 & 0.72 \\
Faster R-CNN        & CPU & FP32 & 1312.9 & 0.8  & 1838 & -- & -- \\
Faster R-CNN        & GPU & FP32 & 68.8   & 14.5 & 708  & 68.2 & 4.69 \\
Faster R-CNN        & GPU & FP16 & 39.0   & 25.7 & 391  & 60.2 & 2.35 \\
OWLv2-Base          & CPU & FP32 & 1649.3 & 0.6  & 1941 & -- & -- \\
OWLv2-Base          & GPU & FP32 & 277.0  & 3.6  & 817  & 54.7 & 15.15 \\
OWLv2-Base          & GPU & FP16 & 166.5  & 6.0  & 426  & 39.6 & 6.59 \\
\bottomrule
\end{tabular}
\end{table}
\FloatBarrier

Table~\ref{tab:efficiency} reports the deployment cost of the three closed-set detectors and OWLv2-Base. These measurements rank the models on a server CPU and a data-center GPU; they are not measurements on the embedded devices of a drone or a sorting-line controller. CPU latency separates the models into three tiers. YOLO26-s (44.7~ms, 22.4~fps) runs in real time without an accelerator, RF-DETR-small (133.5~ms) is three times slower, and Faster R-CNN and OWLv2-Base stay under 1~fps. On the GPU, YOLO26-s also has the smallest peak memory (98~MB) and the lowest energy per image (0.33~J), against 0.60~J for RF-DETR-small, 4.69~J for Faster R-CNN and 15.15~J for OWLv2-Base, a factor of 46 between the cheapest and the most expensive model. The zero-shot ability of OWLv2-Base therefore has a high cost: 24 times the GPU latency of YOLO26-s and 8 times its peak memory (Figure~\ref{fig:efficiency-pareto}). FP16 helps the two largest models (1.76 and 1.66 times faster) but slows down YOLO26-s and RF-DETR-small at batch size 1, and the INT8 YOLO26-s, run with ONNX Runtime, is 3.4 times slower than the FP32 PyTorch model; both results reflect our inference path rather than the potential of reduced precision.

\section{Failure analysis}\label{sec:failure}

\begin{table}[h]
\centering
\caption{Recall at IoU 0.5 (confidence 0.25) by relative-size tercile and scene-density tercile, computed within each dataset, Level 0, test split, using each dataset's best Level 0 model from Table~\ref{tab:s1s3}. Bold marks the values discussed in the text: the smallest-size and densest terciles of TACO, and the densest tercile of WaRP-D, the only reversal of the density effect.}\label{tab:failure-analysis}
\begin{tabular}{@{}lccccccc@{}}
\toprule
Dataset & Overall & \multicolumn{3}{c}{Size (low/mid/high)} & \multicolumn{3}{c}{Density (low/mid/high)} \\
\midrule
SortWaste                    & 0.942 & 0.901 & 0.959 & 0.965 & 0.958 & 0.942 & 0.924 \\
ZeroWaste-f                  & 0.803 & 0.724 & 0.829 & 0.858 & 0.844 & 0.787 & 0.762 \\
WaRP-D                       & 0.841 & 0.799 & 0.865 & 0.859 & 0.737 & 0.890 & \textbf{0.952} \\
TACO                         & 0.475 & \textbf{0.257} & 0.456 & 0.713 & 0.704 & 0.447 & \textbf{0.255} \\
UAVVaste                     & 0.797 & 0.702 & 0.816 & 0.875 & 0.889 & 0.840 & 0.615 \\
IWHR-Floater & 0.865 & 0.780 & 0.895 & 0.919 & 0.936 & 0.909 & 0.736 \\
\bottomrule
\end{tabular}
\end{table}
\FloatBarrier

For each dataset, we take the best Level 0 model of Table~\ref{tab:s1s3} and compute its in-domain recall by size and density tercile (Table~\ref{tab:failure-analysis}). Recall increases with object size on five of the six datasets, WaRP-D being the exception (0.865 for the middle tercile against 0.859 for the largest), and on these five datasets the difference between the largest and smallest terciles ranges from 0.064 on SortWaste to 0.456 on TACO (0.257 to 0.713), where the smallest objects are missed nearly three times as often as they are found (Figure~\ref{fig:taco-size}). Recall decreases with scene density on five datasets, by 0.034 on SortWaste up to 0.274 on UAVVaste and 0.449 on TACO. WaRP-D is the only reversal, from 0.737 in the sparsest tercile to 0.952 in the densest. This is due to the fact that its sparse images often show a few annotated items in a bin full of similar, unannotated material, whereas its dense images can show well-separated items on a clean belt (Figure~\ref{fig:warp-density}). In WaRP-D, the annotated count therefore does not measure visual clutter, which is the annotation-scope effect of Section~\ref{sec:taxonomy}.

\section{Discussion}\label{sec:discussion}

\subsection{Implications for waste management practice}

The results point to different choices for different applications rather than to a single best model. Since all costs were measured on a server CPU and a data-center GPU, the recommendations below rank the models; they do not guarantee a frame rate on a given embedded device.

\textbf{Sorting lines.} Sorting is a material-level task, so the relevant accuracy is Level 1. RF-DETR-small is the strongest in-domain model on the three conveyor datasets and runs at 17.1~ms per image on the GPU, which makes it the natural choice when the line has a GPU. Without a GPU, YOLO26-s is the only model below 50~ms on our CPU, but this costs 0.19 Level 1 mAP@0.5 on SortWaste and gives a metal recall of only 0.123. Whether this is acceptable depends on the belt speed and on the time between capture and separation, which our benchmark does not model. A model moved to a new plant loses most of its accuracy, even when it was trained on other conveyors, so a new line should plan for local labels: with a hundred labeled images, YOLO26-s from its S3 checkpoint reaches 0.611 on SortWaste at Level 1, above its S1 result.

\textbf{Drone and water monitoring.} On UAVVaste, YOLO26-s trained on the five other datasets reaches 0.577 without any aerial training image, 90\% of its in-domain result, and OWLv2-Base reaches 0.685 zero-shot when the images can be processed after the flight on a ground station. On IWHR-Floater, the S3 model reaches only 0.472, and a hundred local training images bring it to 0.796, so a small local labeling effort is worthwhile for a new water site.

\textbf{Litter in uncontrolled scenes.} TACO is the hardest dataset for every detector. A hundred local images bring YOLO26-s from its COCO weights to 0.426 at Level 0, close to the full training split (0.439). Small objects account for most missed detections, so systems that control their capture, such as vehicle-mounted cameras, should keep litter items large in the image through resolution or distance; we did not test this.

\textbf{Detector design.} In-domain accuracy does not predict transfer: RF-DETR-small, whose DINOv2 backbone is pretrained on large-scale data, keeps 48\% of its accuracy on other datasets against 25\% for YOLO26-s, a larger difference than in-domain. Pooling several source datasets helps most the detectors that transfer worst, so multi-source training is a cheap remedy for compact detectors, and an open-vocabulary head is a useful initialization when local labels are scarce.

\subsection{Limitations}

Several limitations of the present evaluation should be kept in mind. First, the three closed-set detectors are not trained under a matched protocol, so the comparisons are between training recipes rather than architectures. Second, only the YOLO26-s S1 and S2 results are repeated over three seeds, and the intervals are computed for YOLO26-s only, so a comparison with another model is checked against the YOLO26-s interval alone. Third, litter, aerial and water are each represented by a single dataset, and the datasets differ in annotation scope, so part of what we measure as a gap between domains is a gap between datasets. Fourth, the few-shot runs select checkpoints and learning rates on the full target validation split, which a new site with only $k$ labeled images would not have. Fifth, the zero-shot prompts were informed by evaluation images, and the VLMs are evaluated on 100-image samples without confidence scores, so they are compared by precision, recall and F1 at a single operating point rather than by mAP. Finally, the deployment measurements cover a single server CPU and a single data-center GPU, and accuracy is not re-evaluated under FP16 or INT8.

\section{Conclusion}\label{sec:conclusion}

This paper presented WasteBench, a benchmark that unifies six public waste detection datasets from four domains under a shared two-level taxonomy, and evaluates closed-set detectors, open-vocabulary detectors and VLMs under five settings, together with their deployment cost. The gap between datasets is severe for every closed-set detector: on average, a model trained on one dataset keeps 18\% to 48\% of its in-domain accuracy on the others, and transfer between conveyor datasets from different plants is only slightly better than transfer across domains. Without target labels, OWLv2-Base is a better starting point than YOLO26-s trained on the other datasets, but not than RF-DETR-small, and the S3 Faster R-CNN reaches a similar mean (0.439 against 0.424); it costs 46 times the GPU energy per image of YOLO26-s. A hundred labeled target images, with a validation split for selection, recover 77\% to 93\% of the in-domain accuracy. Recall falls for small objects and dense scenes, most strongly on TACO. For waste management practice, these results indicate that a detector should be validated on the target site before deployment, and that a hundred local labeled images recover more of the in-domain accuracy than training on the other datasets alone.

As future work, we plan a class-matched reanalysis of the Level 1 transfer results, repeated training runs for all detectors, measurements on embedded hardware with accuracy under quantization, and additional datasets such as PlastOPol and TrashCan.

\section*{CRediT authorship contribution statement}

\textbf{Salah Eddine Bekhouche:} Conceptualization, Software, Investigation, Formal analysis, Data curation, Visualization, Writing -- original draft.
\textbf{Fadi Abdeladhim Zidi:} Methodology, Writing -- original draft.
\textbf{Ahmed Marouane Djouamaa:} Methodology, Writing -- original draft.
\textbf{Abdellah Zakaria Sellam:} Methodology, Writing -- original draft.
\textbf{Fadi Dornaika:} Validation, Writing -- review \& editing.
\textbf{Cosimo Distante:} Validation, Writing -- review \& editing.
\textbf{Abdenour Hadid:} Validation, Writing -- review \& editing.

\section*{Declaration of competing interest}

The authors declare that they have no known competing financial interests or personal relationships that could have appeared to influence the work reported in this paper.

\section*{Funding}

This research did not receive any specific grant from funding agencies in the public, commercial, or not-for-profit sectors.

\section*{Data availability}

The tools that retrieve each dataset from its original public source (Table~\ref{tab:datasets}), together with the training, evaluation and benchmarking code, are available at \url{https://github.com/Bekhouche/WasteBench}, and the taxonomy tables and split files at \url{https://huggingface.co/datasets/Bekhouche/WasteBench}. The six source datasets are already public under the licenses of Table~\ref{tab:datasets}; we do not redistribute their images, and the released scripts download each dataset from its original source and prepare it for the benchmark.

\section*{Declaration of generative AI and AI-assisted technologies in the manuscript preparation process}

During the preparation of this work the authors used Claude (Anthropic) in order to improve the language and readability of the manuscript. After using this tool, the authors reviewed and edited the content as needed and take full responsibility for the content of the publication.

\bibliographystyle{elsarticle-harv}
\bibliography{references}

\clearpage
\appendix
\setcounter{section}{0}
\setcounter{table}{0}
\setcounter{figure}{0}
\renewcommand{\thesection}{\arabic{section}}
\renewcommand{\thetable}{S\arabic{table}}
\renewcommand{\thefigure}{S\arabic{figure}}
\makeatletter
\renewcommand\@seccntformat[1]{Note~\csname the#1\endcsname.\quad}
\makeatother
\begin{center}{\large\bfseries Supplementary material}\end{center}

This appendix contains the supplementary notes, tables and figures. Tables and figures numbered with the prefix S are in this appendix; tables, figures and sections numbered without it are in the main text.

\section{Repeated images within and across datasets}\label{note:dedup}

Following the warning of the ZeroWaste authors that an augmented variant of their dataset pastes TACO objects into conveyor backgrounds, we check for repeated images over all 18,010 images of the six datasets, both across datasets and across the splits of each dataset. Each image receives a 64-bit perceptual hash (pHash, as implemented in the \texttt{imagehash} library), computed from the discrete cosine transform of a downscaled grayscale copy, so that re-encoded or resized copies of an image usually receive the same hash although their files differ. This hash describes the appearance of the image, not its content: two different photographs of similar objects receive different hashes. Images with identical hashes, that is, at a Hamming distance of zero, are put in the same group, so a group is the set of all images that share one hash and can contain two or more images.

No group contains images from two different datasets, so the six datasets do not overlap. Within datasets, 98 groups occur, with 230 images in total: 85 groups are pairs and 13 contain three to ten images. A group crosses a split boundary when its images do not all belong to the same split. This is the case for 49 of the 98 groups, with 126 images: 37 in IWHR-Floater, 9 in WaRP-D and 3 in ZeroWaste-f, the latter two within the official splits, which we keep for comparability with the original papers. The other 49 groups lie within a single split, where they repeat an image but do not leak it between training and test. Removing the 26 IWHR-Floater test images with an identical hash in its training or validation split changes the Level 0 mAP@0.5 of YOLO26-s and RF-DETR-small from 0.914 to 0.926, so these duplicates do not inflate the score. This test only covers identical hashes; frames of the same fixed camera that are not identical can still fall on both sides of the IWHR-Floater split, so its in-domain results may be optimistic.

\section{Prompts and model interfaces}\label{note:prompts}

The open-vocabulary detectors and the vision-language models differ in their inputs, their outputs and therefore their evaluation settings (Table~\ref{tab:family-interface}). An open-vocabulary detector takes an image and a list of text prompts, encodes all prompts in a single forward pass, and returns boxes with a confidence score for each prompt; several synonyms can therefore be given for one class, and their boxes are mapped back to that class. A vision-language model generates text: it takes an image and a natural-language instruction, and writes the boxes as coordinates in its answer, without a confidence score. Each query requires a separate generation, so we give these models one phrase per class. Florence-2-Base receives its open-vocabulary detection task token, \texttt{<OPEN\_VOCABULARY\_DETECTION>}, followed by the phrase, and Qwen3-VL-2B receives an instruction that asks for a JSON list of boxes, or an empty list when the queried object is absent (Tables~\ref{tab:prompts} and~\ref{tab:prompt-inputs}).

For YOLOE, we use the Ultralytics checkpoint built on YOLO26-s (\texttt{yoloe-26s-seg}) and keep only its boxes. Of its three prompt types, we use the text prompts: each phrase is encoded by its text encoder, and the resulting embeddings replace the classifier of its detection head. Visual prompts would require example boxes from the target dataset, which S4 does not allow, and the prompt-free mode returns names from its own vocabulary of 4,585 categories rather than our classes.

In preliminary experiments with YOLOE, concrete object descriptions such as ``plastic bottle'' produced more detections than the literal taxonomy labels, so we use such concrete-noun prompts for the main comparison and keep the literal-label run of YOLOE as a prompt-sensitivity result. These prompts were then fixed for Grounding DINO-Tiny and OWLv2-Base. They were informed by examining evaluation images, although no model weights were fitted to WasteBench data, and they omit the Level 1 \textit{other} class (120 of the 548 TACO instances), whose instances are excluded from the corresponding evaluations; the comparison between the two prompt sets at Level 1 is therefore not a prompt-only one. The open-vocabulary detectors keep their boxes above a confidence of 0.1 and are evaluated on the complete test splits. The vision-language models are evaluated on a fixed, seeded sample of 100 test images per dataset (the full split when it is smaller).

\section{\texorpdfstring{Evaluation of models without confidence scores}{Evaluation of models without confidence scores}}\label{note:vlm-ap}

AP is computed per class by sorting the boxes by their confidence score, matching them greedily to ground-truth boxes at IoU 0.5, and interpolating the precision-recall curve accumulated along this order. The vision-language models return boxes without a confidence score, so their boxes have no ranking. Any score given to them, such as 1.0 for every box, leaves the boxes in an arbitrary order (the order of the images and, within an image, the order of generation), and the resulting AP depends on this order rather than on the model. We therefore do not report mAP for the vision-language models. Their output is a single operating point, which we evaluate by precision, recall and F1 at IoU 0.5, keeping every box they return. For a like-for-like comparison, the open-vocabulary detectors are evaluated in the same way on the same images, at a confidence threshold of 0.25 (Table~\ref{tab:s4-matched}). On these samples, the mAP of the detectors is close to their full-split values (0.426 against 0.424 for OWLv2-Base at Level 0), and YOLOE has a high precision but a very low recall, since most of its zero-shot scores fall below the 0.25 threshold.

\section{Deployment measurements}\label{note:deployment}

We measure median latency, throughput, peak memory and GPU power at batch size 1 for the three closed-set detectors and OWLv2-Base, the strongest zero-shot detector, on an Intel Xeon Silver 4314 server (two sockets, 32 cores) and a single NVIDIA L4 GPU, in FP32 and, on the GPU, FP16. All models run in PyTorch with its default of 32 CPU threads, on a host shared with other jobs. Each configuration is timed over 50 runs after five warm-up runs on one decoded SortWaste test image; the timed step covers preprocessing, the forward pass and post-processing (box decoding and thresholding). Each model uses its default inference resolution: 640 pixels for YOLO26-s, a shorter side of 800 pixels (longer side up to 1,333) for the torchvision Faster R-CNN, $960 \times 960$ for OWLv2-Base, and the released default for RF-DETR-small. GPU power is the board power reported by \texttt{nvidia-smi}, sampled every 0.1~s during the timed runs and averaged; energy per image is this mean power times the median latency, and includes the idle power of the board. CPU power is not measured. We also time an INT8 post-training-quantized YOLO26-s, calibrated on SortWaste validation images and run with ONNX Runtime on the CPU. Accuracy is not re-evaluated under FP16 or INT8.

FP16 helps the two largest models, Faster R-CNN and OWLv2-Base (1.76 and 1.66 times faster, with about half the memory), but slows down YOLO26-s and RF-DETR-small by 51\% and 23\% at batch size 1. We did not profile this effect; it is likely specific to our plain PyTorch inference path, in which casting and kernel launches weigh more than the arithmetic for a small model, and compiled runtimes such as TensorRT may behave differently. Likewise, the INT8 YOLO26-s is 3.4 times slower than the FP32 PyTorch model on the same CPU (153.7~ms); since the two use different runtimes, this result reflects our INT8 deployment path rather than the potential of INT8 quantization.

\section{Source and target datasets under pairwise transfer}\label{note:s2-patterns}

Single S2 pairs vary more between training seeds than their means, by up to 0.13 mAP@0.5 (SortWaste to IWHR-Floater at Level 0), so we interpret the means rather than individual pairs (Table~\ref{tab:s2-summary}). Several patterns hold for all three detectors: SortWaste is the easiest target, ZeroWaste-f is the hardest (for YOLO26-s, joint hardest) target at Level 0, and models trained on UAVVaste transfer worst to the conveyor datasets. WaRP-D is a hard target although it is fitted well in-domain (0.732 at Level 0). On the source side, TACO, the weakest in-domain dataset, is the best Level 0 source for YOLO26-s (0.381, 1.7 times the next-best, IWHR-Floater; it is the better of the two in all 1,000 bootstrap resamples) and for Faster R-CNN (0.452), whereas SortWaste is one of the two worst YOLO26-s sources. This may be due to the diversity of the crowdsourced scenes of TACO against the consistency of a single conveyor, but it is partly specific to the training recipe: for RF-DETR-small, the six source means lie within 0.325 to 0.392, and SortWaste is a mid-ranked source (0.379).

\section{Few-shot starting points and model selection}\label{note:fewshot}

YOLOE starts S5 from the text embeddings of the class names (``trash'' at Level 0; ``plastic'', ``paper'', ``metal'', ``glass'' and ``trash'' at Level 1), so that its curve starts from the zero-shot model. Its $k=0$ point uses a single prompt per class and the same evaluator as the fine-tuned runs, which explains why it differs from the S4 values, obtained with several synonyms per class and the COCO evaluation protocol. In all three cases, $k=0$ means that no labeled image of the target dataset is used, not that the model is zero-shot. Only YOLOE is zero-shot at $k=0$, since it has seen no WasteBench label. YOLO26-s is a closed-set detector: its $k=0$ point is the S3 result, obtained after training with the same classes on the other datasets, so it has seen labeled waste, but no image of the target dataset.

Every run uses up to 100 epochs with an early-stopping patience of 30 and selects its checkpoint on the full target validation split. Each configuration is trained at the default learning rate and at one tenth of it with AdamW, and we keep the rate with the higher validation mAP@[0.5:0.95] (the lower rate in 82 of the 90 configurations). The labeled target data seen by a run is therefore $k$ training images plus the validation split, 94 to 780 images depending on the dataset; the validation split is used only for selection, not for gradient updates, but a site with only $k$ labeled images would have to select checkpoints on fewer of them. This learning-rate search is not applied to the S1 baseline. On UAVVaste, the COCO start reaches 0.716 at $k=50$, within the S1 interval (0.558 to 0.735), and drops to 0.684 at $k=100$, so differences of a few hundredths of mAP@0.5 between neighboring budgets are within training noise.

\clearpage

\section*{Evaluation protocol and models}

\begin{table}[h]
\centering
\caption{The five evaluation settings.}\label{tab:settings}
\begin{tabular}{@{}p{2.6cm}p{2.8cm}p{2.3cm}p{4cm}@{}}
\toprule
Setting & Train & Test & Question \\
\midrule
S1, in-domain              & Dataset X train          & Dataset X test & In-domain reference \\
S2, cross-domain (pairwise) & Dataset X train         & Dataset Y test & Size of the gap between datasets (RQ1) \\
S3, leave-one-domain-out    & Train splits of all datasets except Y & Dataset Y test & Generalization to an unseen domain (RQ1) \\
S4, zero-shot               & None (pretrained only)  & All test splits  & Open-vocabulary and VLM transfer (RQ2) \\
S5, few-shot                & $k=10,50,100$ training images from Y, starting from S3 YOLO26-s, COCO YOLO26-s or YOLOE; full Y validation split for checkpoint selection & Dataset Y test & Label efficiency on a new site, by starting point (RQ2) \\
\bottomrule
\end{tabular}
\end{table}
\FloatBarrier

\begin{table}[h]
\centering
\caption{Model families and the settings each is evaluated under.}\label{tab:models}
\begin{tabular}{@{}lll@{}}
\toprule
Family & Models & Settings \\
\midrule
Closed-set detectors        & YOLO26-s                                & S1, S2, S3, S5 \\
Closed-set detectors        & RF-DETR-small                           & S1, S2, S3 \\
Closed-set detectors        & Faster R-CNN                            & S1, S2, S3 (Level 0) \\
Open-vocabulary detectors   & YOLOE                                   & S4, S5 \\
Open-vocabulary detectors   & Grounding DINO-Tiny, OWLv2-Base                   & S4 \\
Vision-language models      & Florence-2-Base, Qwen3-VL-2B                  & S4 (as grounding detectors) \\
\bottomrule
\end{tabular}
\end{table}
\FloatBarrier

\begin{table}[h]
\centering
\caption{Inputs, outputs and evaluation settings of the three model families. Under S4, the open-vocabulary detectors and the vision-language models receive the prompts of Table~\ref{tab:prompts}.}\label{tab:family-interface}
\footnotesize
\setlength{\tabcolsep}{4pt}
\begin{tabular}{@{}p{2.3cm}p{2.9cm}p{3.1cm}p{3.1cm}@{}}
\toprule
 & Closed-set detector & Open-vocabulary detector & Vision-language model \\
\midrule
Input             & Image & Image and a list of text prompts & Image and a natural-language instruction \\
Classes           & Fixed at training (Level 0 or Level 1) & Any class named in the prompts & Any class named in the instruction \\
Output            & Boxes with a class and a confidence score & Boxes with a confidence score per prompt & Box coordinates written in the generated text, without a score \\
Calls per image   & One forward pass & One forward pass for all prompts & One generation per class phrase \\
Prompts per class & -- & Several synonyms, mapped back to the class & One phrase \\
Test images (S4)  & -- & Full test splits & Up to 100 per dataset (fixed sample) \\
Box score         & Model score & Model score, boxes above 0.1 & 1.0 for every box \\
Settings          & S1, S2, S3, S5 & S4; S5 for YOLOE & S4 \\
\bottomrule
\end{tabular}
\end{table}
\FloatBarrier

\begin{table}[h]
\centering
\caption{Zero-shot text prompts per class (S4). The open-vocabulary detectors receive all phrases of a class in one forward pass, and their boxes are mapped back to that class; the vision-language models receive one phrase per class, one generation per phrase. Table~\ref{tab:prompt-inputs} shows the exact text each model receives. The literal-label run of YOLOE uses the taxonomy labels instead, including \textit{other}. The classes are those of the WasteBench taxonomy, not the native classes of each dataset (8 for SortWaste, 28 for WaRP-D and 60 for TACO; Table~\ref{tab:datasets}): every native class is first mapped to one Level 1 material (Section~\ref{sec:taxonomy}), and the predicted boxes are scored against the mapped labels. The four Level 1 rows are therefore materials shared by all datasets, not superclasses of one dataset. The fifth material, \textit{other}, has no prompt, and its instances are excluded from the Level 1 zero-shot evaluation (Note~\ref{note:prompts}).}\label{tab:prompts}
\footnotesize
\setlength{\tabcolsep}{4pt}
\begin{tabular}{@{}llp{4.8cm}ll@{}}
\toprule
Level & Class & Open-vocabulary detectors & Florence-2-Base & Qwen3-VL-2B \\
\midrule
0 & waste           & trash, litter, garbage, debris, plastic bottle, cardboard box, metal can, glass bottle & trash & trash or litter \\
1 & plastic         & plastic, plastic bottle, plastic bag & plastic   & plastic \\
1 & paper/cardboard & cardboard, cardboard box, paper      & cardboard & cardboard \\
1 & metal           & metal, metal can, tin can            & metal     & metal \\
1 & glass           & glass, glass bottle                  & glass     & glass \\
\bottomrule
\end{tabular}
\end{table}
\FloatBarrier

\begin{table}[h]
\centering
\caption{Exact text input of each zero-shot model (S4), shown for Level 0. At Level 1, the phrases of Table~\ref{tab:prompts} replace those of the waste class in the same format. The same input is used for every image of every dataset.}\label{tab:prompt-inputs}
\footnotesize
\begin{tabular}{@{}p{2.6cm}p{2.2cm}p{7cm}@{}}
\toprule
Model & Calls per image & Text input \\
\midrule
Grounding DINO-Tiny & One & \texttt{trash. litter. garbage. debris. plastic bottle. cardboard box. metal can. glass bottle.} \\
OWLv2-Base, YOLOE   & One & The list [\texttt{trash}, \texttt{litter}, \texttt{garbage}, \texttt{debris}, \texttt{plastic bottle}, \texttt{cardboard box}, \texttt{metal can}, \texttt{glass bottle}], one text embedding per phrase \\
Florence-2-Base     & One per class & \texttt{<OPEN\_VOCABULARY\_DETECTION>trash} \\
Qwen3-VL-2B         & One per class & Detect all instances of ``trash or litter'' in the image. If none are present, output an empty JSON list []. Otherwise output a JSON list of objects, each with a bbox\_2d field [x1,y1,x2,y2] and a label field. \\
\bottomrule
\end{tabular}
\end{table}
\FloatBarrier

\section*{In-domain results (S1)}

\begin{table}[h]
\centering
\caption{YOLO26-s, in-domain, Level 1, test split, all four datasets with a material signal.}\label{tab:s1-summary}
\begin{tabular}{@{}lccccc@{}}
\toprule
Dataset & Instances & Precision & Recall & mAP@0.5 & mAP@[0.5:0.95] \\
\midrule
SortWaste   & 12,618 & 0.555 & 0.539 & 0.530 & 0.398 \\
ZeroWaste-f & 5,077  & 0.501 & 0.423 & 0.404 & 0.287 \\
WaRP-D      & 1,551  & 0.707 & 0.500 & 0.564 & 0.403 \\
TACO        & 548    & 0.471 & 0.222 & 0.164 & 0.115 \\
\bottomrule
\end{tabular}
\end{table}
\FloatBarrier

\begin{table}[h]
\centering
\caption{YOLO26-s, in-domain, SortWaste test split (776 images, 12,618 objects).}\label{tab:s1-sortwaste}
\begin{tabular}{@{}lccccc@{}}
\toprule
Class & Instances & Precision & Recall & mAP@0.5 & mAP@[0.5:0.95] \\
\midrule
plastic         & 9,170  & 0.665 & 0.807 & 0.770 & 0.573 \\
paper/cardboard  & 3,233  & 0.608 & 0.686 & 0.653 & 0.495 \\
metal           & 215    & 0.392 & 0.123 & 0.168 & 0.125 \\
\bottomrule
\end{tabular}
\end{table}
\FloatBarrier

\begin{table}[h]
\centering
\caption{YOLO26-s, in-domain, Level 0, test split, all six datasets.}\label{tab:s1-level0}
\begin{tabular}{@{}lccccc@{}}
\toprule
Dataset & Instances & Precision & Recall & mAP@0.5 & mAP@[0.5:0.95] \\
\midrule
SortWaste                       & 12,618 & 0.810 & 0.809 & 0.858 & 0.643 \\
ZeroWaste-f                     & 5,077  & 0.591 & 0.594 & 0.567 & 0.391 \\
WaRP-D                          & 1,551  & 0.741 & 0.657 & 0.732 & 0.552 \\
TACO                            & 651    & 0.600 & 0.429 & 0.439 & 0.301 \\
UAVVaste                        & 340    & 0.734 & 0.579 & 0.640 & 0.326 \\
IWHR-Floater    & 3,586  & 0.904 & 0.823 & 0.914 & 0.673 \\
\bottomrule
\end{tabular}
\end{table}
\FloatBarrier

\begin{table}[h]
\centering
\caption{Uncertainty of the YOLO26-s results of Table~\ref{tab:s1s3}, mAP@0.5, test split. The 95\% intervals are estimated from 1,000 bootstrap resamples of the test images (Section~\ref{sec:metrics}); ``3 seeds'' is the mean and standard deviation of S1 over three training seeds (Section~\ref{sec:implementation}). S3 is trained with a single seed.}\label{tab:yolo-uncertainty}
\footnotesize
\setlength{\tabcolsep}{4pt}
\begin{tabular}{@{}llccccc@{}}
\toprule
Dataset & Level & S1 & S1 95\% interval & S1, 3 seeds & S3 & S3 95\% interval \\
\midrule
SortWaste    & 1 & 0.530 & [0.516, 0.552] & 0.532 $\pm$ 0.003 & 0.304 & [0.294, 0.322] \\
ZeroWaste-f  & 1 & 0.404 & [0.375, 0.444] & 0.440 $\pm$ 0.034 & 0.094 & [0.071, 0.125] \\
WaRP-D       & 1 & 0.564 & [0.520, 0.605] & 0.563 $\pm$ 0.010 & 0.086 & [0.077, 0.106] \\
TACO         & 1 & 0.164 & [0.116, 0.217] & 0.178 $\pm$ 0.012 & 0.026 & [0.013, 0.049] \\
SortWaste    & 0 & 0.858 & [0.852, 0.864] & 0.861 $\pm$ 0.005 & 0.696 & [0.687, 0.705] \\
ZeroWaste-f  & 0 & 0.567 & [0.551, 0.584] & 0.561 $\pm$ 0.015 & 0.138 & [0.127, 0.149] \\
WaRP-D       & 0 & 0.732 & [0.700, 0.759] & 0.730 $\pm$ 0.008 & 0.105 & [0.092, 0.124] \\
TACO         & 0 & 0.439 & [0.381, 0.504] & 0.447 $\pm$ 0.007 & 0.219 & [0.173, 0.279] \\
UAVVaste     & 0 & 0.640 & [0.558, 0.735] & 0.630 $\pm$ 0.012 & 0.577 & [0.507, 0.655] \\
IWHR-Floater & 0 & 0.914 & [0.891, 0.932] & 0.916 $\pm$ 0.002 & 0.472 & [0.428, 0.514] \\
\bottomrule
\end{tabular}
\end{table}
\FloatBarrier

\section*{Pairwise transfer (S2)}

\begin{table}[h]
\centering
\caption{S2 cross-domain transfer, YOLO26-s, mean mAP@0.5 of each dataset as a source (training) dataset and as a target (test) dataset, averaged over the other datasets at the same level, with 95\% bootstrap intervals in brackets.}\label{tab:s2-summary}
\begin{tabular}{@{}llcc@{}}
\toprule
Dataset & Level & Mean as source & Mean as target \\
\midrule
SortWaste                    & 1 & 0.038 [0.034, 0.048] & 0.180 [0.177, 0.185] \\
ZeroWaste-f                  & 1 & 0.096 [0.092, 0.105] & 0.067 [0.053, 0.089] \\
WaRP-D                       & 1 & 0.072 [0.057, 0.091] & 0.035 [0.032, 0.044] \\
TACO                         & 1 & 0.099 [0.093, 0.110] & 0.022 [0.011, 0.036] \\
SortWaste                    & 0 & 0.068 [0.063, 0.076] & 0.440 [0.434, 0.448] \\
ZeroWaste-f                  & 0 & 0.181 [0.174, 0.191] & 0.062 [0.057, 0.066] \\
WaRP-D                       & 0 & 0.128 [0.120, 0.139] & 0.062 [0.055, 0.070] \\
TACO                         & 0 & 0.381 [0.364, 0.402] & 0.071 [0.053, 0.099] \\
UAVVaste                     & 0 & 0.059 [0.052, 0.068] & 0.204 [0.181, 0.233] \\
IWHR-Floater & 0 & 0.223 [0.208, 0.242] & 0.202 [0.184, 0.223] \\
\bottomrule
\end{tabular}
\end{table}
\FloatBarrier

\begin{table}[h]
\centering
\caption{S2 cross-domain transfer for the three closed-set detectors, mAP@0.5, test split. In-domain is the mean S1 result over the datasets of each level; cross-domain is the mean over all source-target pairs (12 at Level 1, 30 at Level 0); Kept is their ratio. For YOLO26-s, the 95\% bootstrap interval of Kept is 17.4 to 19.8\% at Level 1 and 24.5 to 25.9\% at Level 0.}\label{tab:s2-models}
\footnotesize
\setlength{\tabcolsep}{4pt}
\begin{tabular}{@{}lcccccc@{}}
\toprule
 & \multicolumn{3}{c}{Level 1} & \multicolumn{3}{c}{Level 0} \\
Model & In-domain & Cross-domain & Kept & In-domain & Cross-domain & Kept \\
\midrule
YOLO26-s      & 0.416 & 0.076 & 18\% & 0.692 & 0.173 & 25\% \\
Faster R-CNN  & 0.449 & 0.091 & 20\% & 0.735 & 0.276 & 38\% \\
RF-DETR-small & 0.591 & 0.172 & 29\% & 0.771 & 0.370 & 48\% \\
\bottomrule
\end{tabular}
\end{table}
\FloatBarrier

\begin{table}[h]
\centering
\caption{S2 transfer within the conveyor domain (the six ordered pairs of SortWaste, ZeroWaste-f and WaRP-D) and across domains (all other pairs: 6 at Level 1, 24 at Level 0), mean mAP@0.5. ``Conveyor S1'' is the mean in-domain result over the three conveyor datasets. 95\% bootstrap intervals for YOLO26-s in brackets.}\label{tab:s2-within}
\footnotesize
\setlength{\tabcolsep}{3pt}
\begin{tabular}{@{}lcccccc@{}}
\toprule
 & \multicolumn{3}{c}{Level 1} & \multicolumn{3}{c}{Level 0} \\
Model & Conveyor S1 & Within & Across & Conveyor S1 & Within & Across \\
\midrule
YOLO26-s      & 0.499 & 0.092 & 0.060 & 0.719 & 0.221 & 0.162 \\
              &       & [0.086, 0.101] & [0.054, 0.070] & & [0.218, 0.227] & [0.155, 0.172] \\
Faster R-CNN  & 0.535 & 0.111 & 0.070 & 0.759 & 0.301 & 0.270 \\
RF-DETR-small & 0.671 & 0.226 & 0.118 & 0.805 & 0.377 & 0.368 \\
\bottomrule
\end{tabular}
\end{table}
\FloatBarrier

\section*{Zero-shot and few-shot results (S4 and S5)}

\begin{table}[h]
\centering
\caption{S4 on a matched sample: every model on the same fixed sample of up to 100 test images per dataset, mean over the datasets of each level. mAP@0.5 as in Table~\ref{tab:s4-summary} for the open-vocabulary detectors only, since the VLMs give no confidence score (Section~\ref{sec:metrics} and Note~\ref{note:vlm-ap}); the models are compared by F1. Precision (P), recall (R) and F1 at IoU 0.5; open-vocabulary detectors at a confidence threshold of 0.25, VLMs with all their boxes.}\label{tab:s4-matched}
\footnotesize
\setlength{\tabcolsep}{4pt}
\begin{tabular}{@{}lcccccccc@{}}
\toprule
 & \multicolumn{4}{c}{Level 0} & \multicolumn{4}{c}{Level 1} \\
Model & mAP@0.5 & P & R & F1 & mAP@0.5 & P & R & F1 \\
\midrule
YOLOE               & 0.073 & 0.615 & 0.026 & 0.049 & 0.076 & 0.480 & 0.042 & 0.075 \\
Grounding DINO-Tiny & 0.296 & 0.308 & 0.386 & 0.313 & 0.144 & 0.226 & 0.365 & 0.241 \\
OWLv2-Base          & \textbf{0.426} & 0.444 & 0.534 & \textbf{0.452} & \textbf{0.207} & 0.389 & 0.438 & \textbf{0.363} \\
Florence-2-Base     & -- & 0.337 & 0.120 & 0.172 & -- & 0.204 & 0.332 & 0.230 \\
Qwen3-VL-2B         & -- & 0.393 & 0.195 & 0.249 & -- & 0.330 & 0.383 & 0.326 \\
\bottomrule
\end{tabular}
\end{table}
\FloatBarrier

\begin{sidewaystable}[h]
\centering
\caption{S5 few-shot fine-tuning from three starting points, mAP@0.5, test split. $k=0$ means no labeled target image. For the S3 start, $k=0$ is the S3 result (Table~\ref{tab:s1s3}), not a zero-shot result, since YOLO26-s is trained on the other datasets; for YOLOE, $k=0$ is its zero-shot result with the S5 prompts and evaluation. The COCO start has no $k=0$ result, since its detection head is new. Bold marks the best starting point for each dataset, level and $k \geq 10$.}\label{tab:s5-results}
\footnotesize
\setlength{\tabcolsep}{4pt}
\begin{tabular}{@{}llc|cccc|ccc|cccc@{}}
\toprule
 & & & \multicolumn{4}{c|}{YOLO26-s from S3} & \multicolumn{3}{c|}{YOLO26-s from COCO} & \multicolumn{4}{c}{YOLOE} \\
Dataset & Level & S1 & $k$=0 & 10 & 50 & 100 & 10 & 50 & 100 & $k$=0 & 10 & 50 & 100 \\
\midrule
SortWaste   & 1 & 0.530 & 0.304 & \textbf{0.467} & \textbf{0.573} & \textbf{0.611} & 0.311 & 0.475 & 0.519 & 0.213 & 0.369 & 0.492 & 0.528 \\
ZeroWaste-f & 1 & 0.404 & 0.094 & \textbf{0.178} & \textbf{0.333} & \textbf{0.371} & 0.088 & 0.277 & 0.345 & 0.063 & 0.177 & 0.281 & 0.316 \\
WaRP-D      & 1 & 0.564 & 0.086 & \textbf{0.116} & \textbf{0.206} & \textbf{0.272} & 0.006 & 0.185 & 0.263 & 0.038 & 0.073 & 0.174 & 0.242 \\
TACO        & 1 & 0.164 & 0.026 & 0.034 & 0.076 & 0.095 & 0.016 & 0.105 & 0.126 & 0.087 & \textbf{0.106} & \textbf{0.136} & \textbf{0.168} \\
SortWaste   & 0 & 0.858 & 0.696 & \textbf{0.825} & \textbf{0.858} & 0.865 & 0.781 & 0.848 & \textbf{0.869} & 0.323 & 0.814 & 0.849 & 0.862 \\
ZeroWaste-f & 0 & 0.567 & 0.138 & 0.302 & 0.395 & 0.434 & 0.294 & 0.431 & 0.476 & 0.109 & \textbf{0.353} & \textbf{0.456} & \textbf{0.492} \\
WaRP-D      & 0 & 0.732 & 0.105 & 0.161 & 0.417 & 0.511 & 0.258 & \textbf{0.468} & \textbf{0.570} & 0.031 & \textbf{0.322} & 0.431 & 0.535 \\
TACO        & 0 & 0.439 & 0.219 & 0.220 & 0.311 & 0.355 & \textbf{0.285} & \textbf{0.377} & \textbf{0.426} & 0.154 & 0.284 & 0.335 & 0.392 \\
UAVVaste    & 0 & 0.640 & 0.577 & \textbf{0.646} & 0.617 & 0.619 & 0.584 & \textbf{0.716} & \textbf{0.684} & 0.303 & 0.558 & 0.656 & 0.642 \\
IWHR-Floater & 0 & 0.914 & 0.472 & 0.662 & 0.763 & 0.796 & 0.662 & 0.755 & \textbf{0.818} & 0.270 & \textbf{0.708} & \textbf{0.765} & 0.812 \\
\bottomrule
\end{tabular}
\end{sidewaystable}
\FloatBarrier

\section*{Deployment cost and failure examples}

\begin{figure}[h]
\centering
\includegraphics[width=0.75\textwidth]{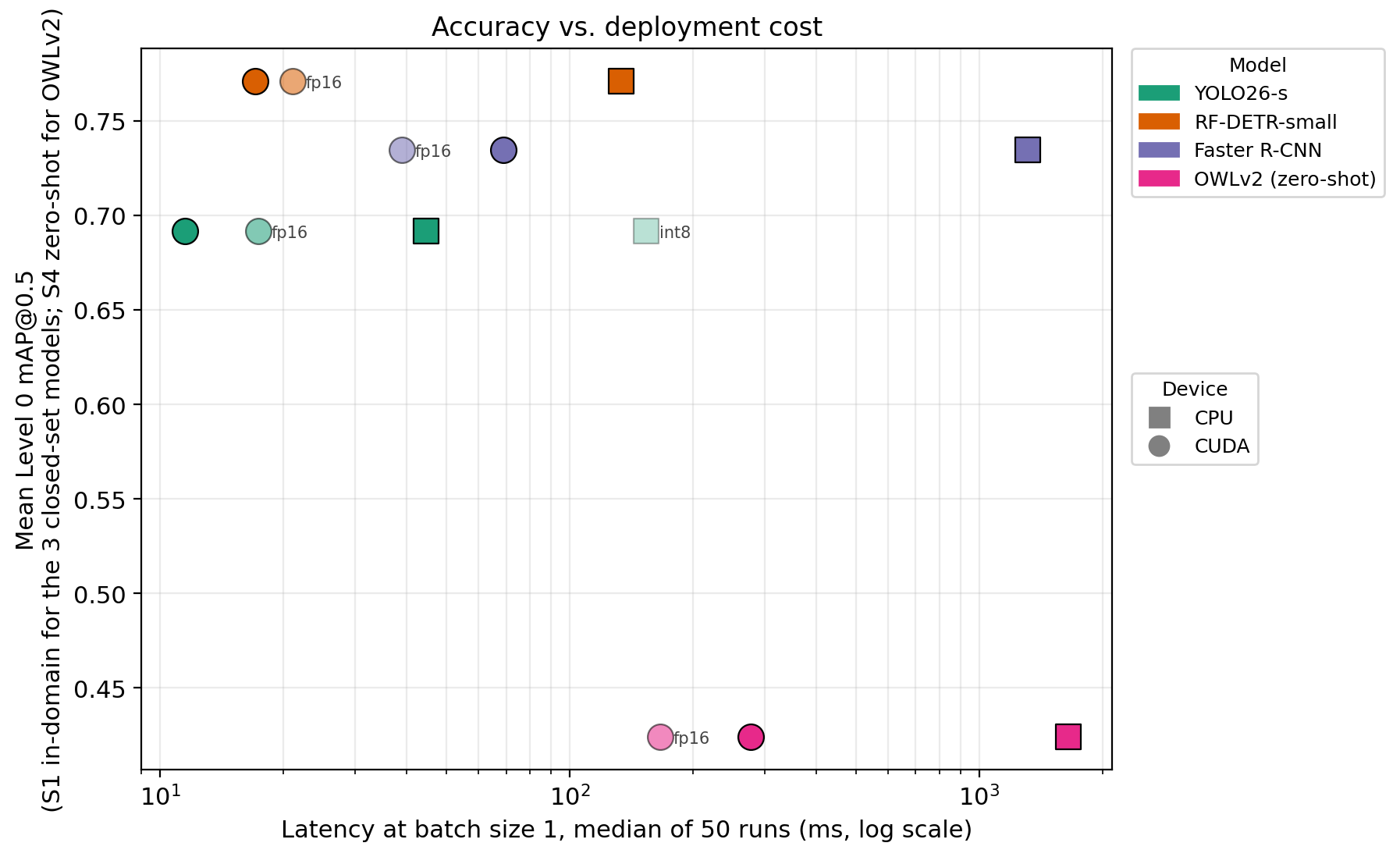}
\caption{Accuracy versus deployment cost. Latency is the median of 50 runs at batch size 1 (Table~\ref{tab:efficiency}); accuracy is each closed-set detector's mean Level 0 in-domain mAP@0.5 (S1 in Table~\ref{tab:s1s3}) or, for OWLv2-Base, its mean Level 0 zero-shot mAP@0.5 (Table~\ref{tab:s4-summary}). The same accuracy value is plotted for every precision of a given model, since only latency and memory, not accuracy, were re-measured per precision.}\label{fig:efficiency-pareto}
\end{figure}
\FloatBarrier

\begin{figure}[h]
\centering
\includegraphics[width=0.48\textwidth]{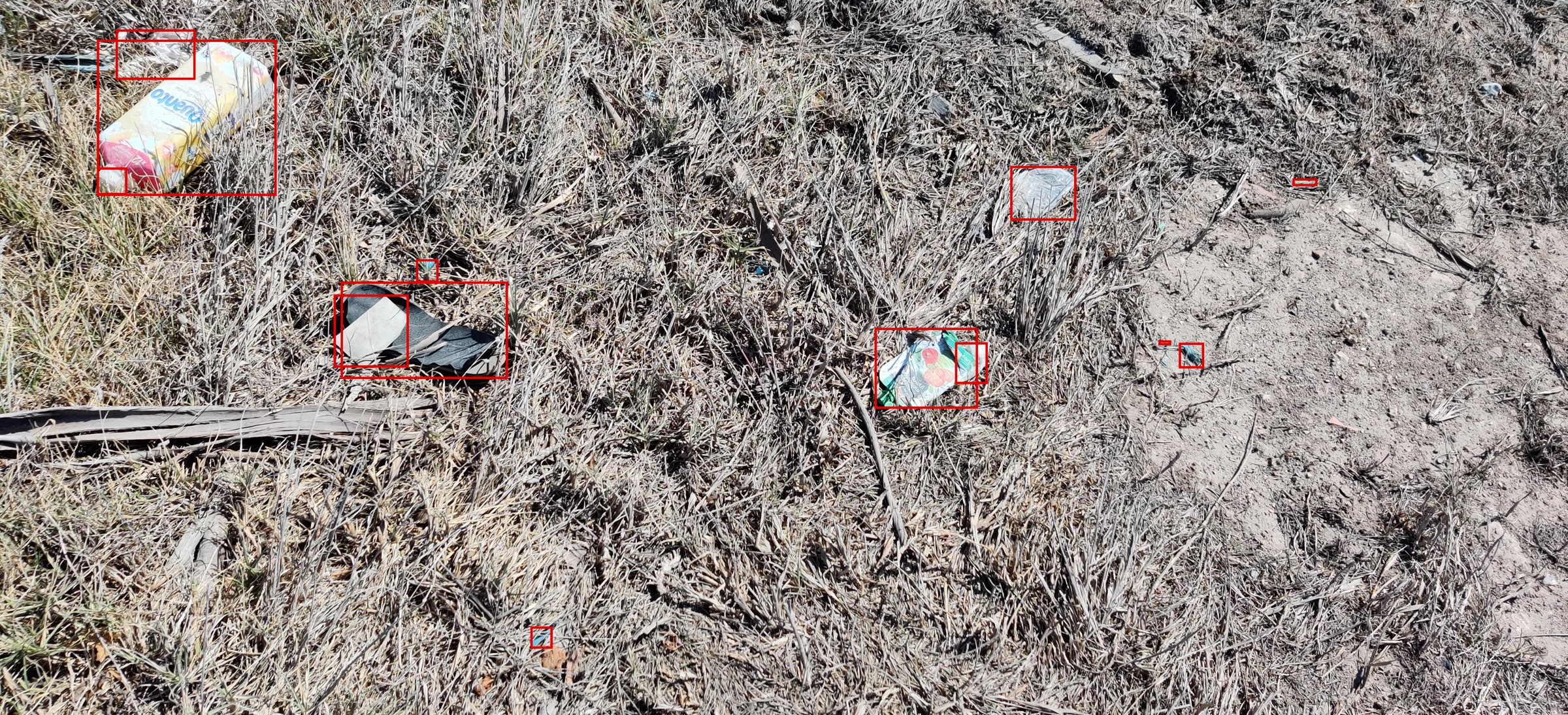}
\hfill
\includegraphics[width=0.48\textwidth]{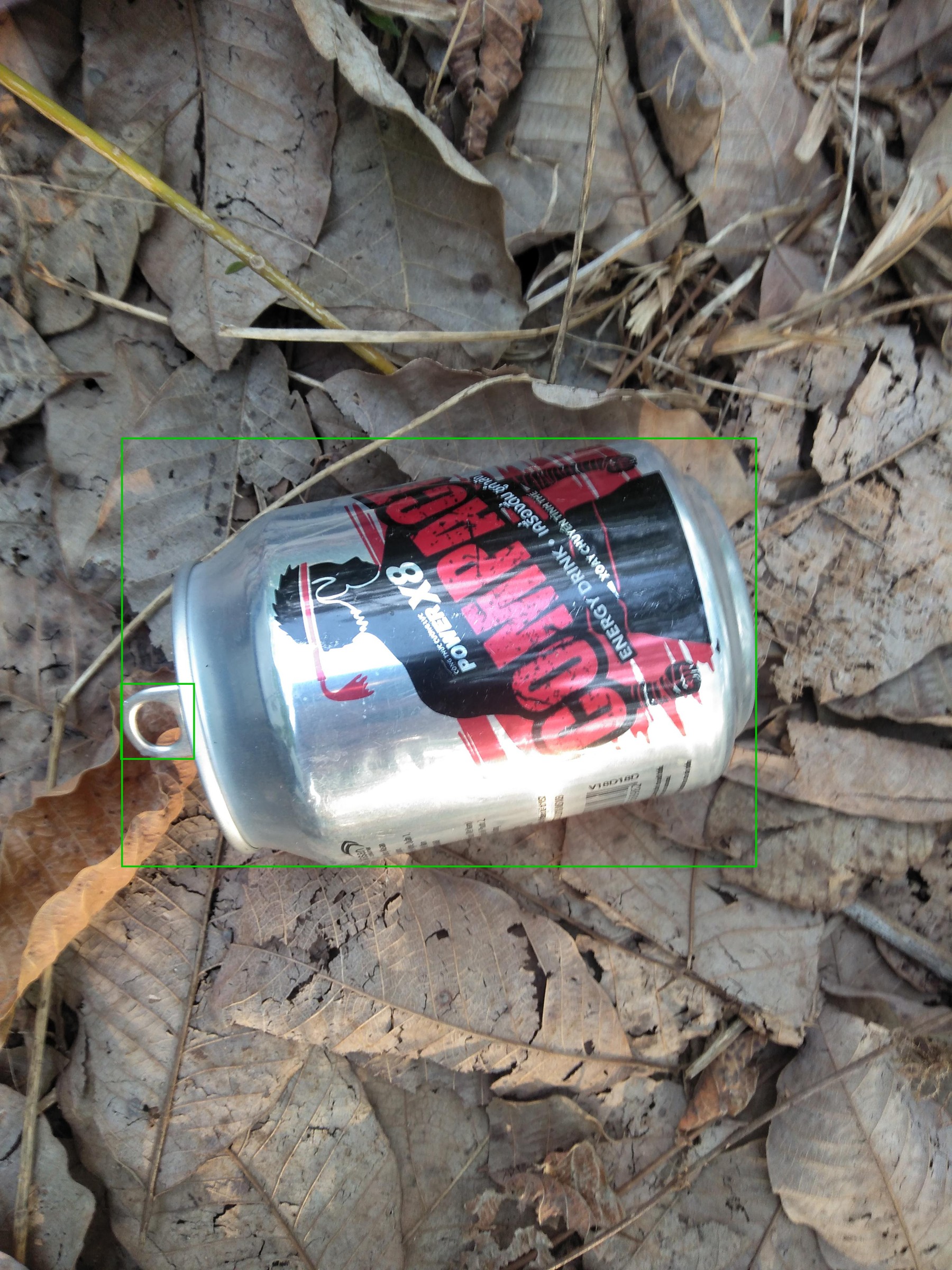}
\caption{TACO, the dataset with the largest size-recall gap in Table~\ref{tab:failure-analysis}. Left: thirteen annotated items on dry grass, mostly small fragments (red boxes), none of them detected, including the two larger cartons. Right: a drink can and its pull tab on dead leaves (green boxes), both detected. The two images also differ in background and clutter, so object size is not the only factor.}\label{fig:taco-size}
\end{figure}
\FloatBarrier

\begin{figure}[h]
\centering
\includegraphics[width=0.48\textwidth]{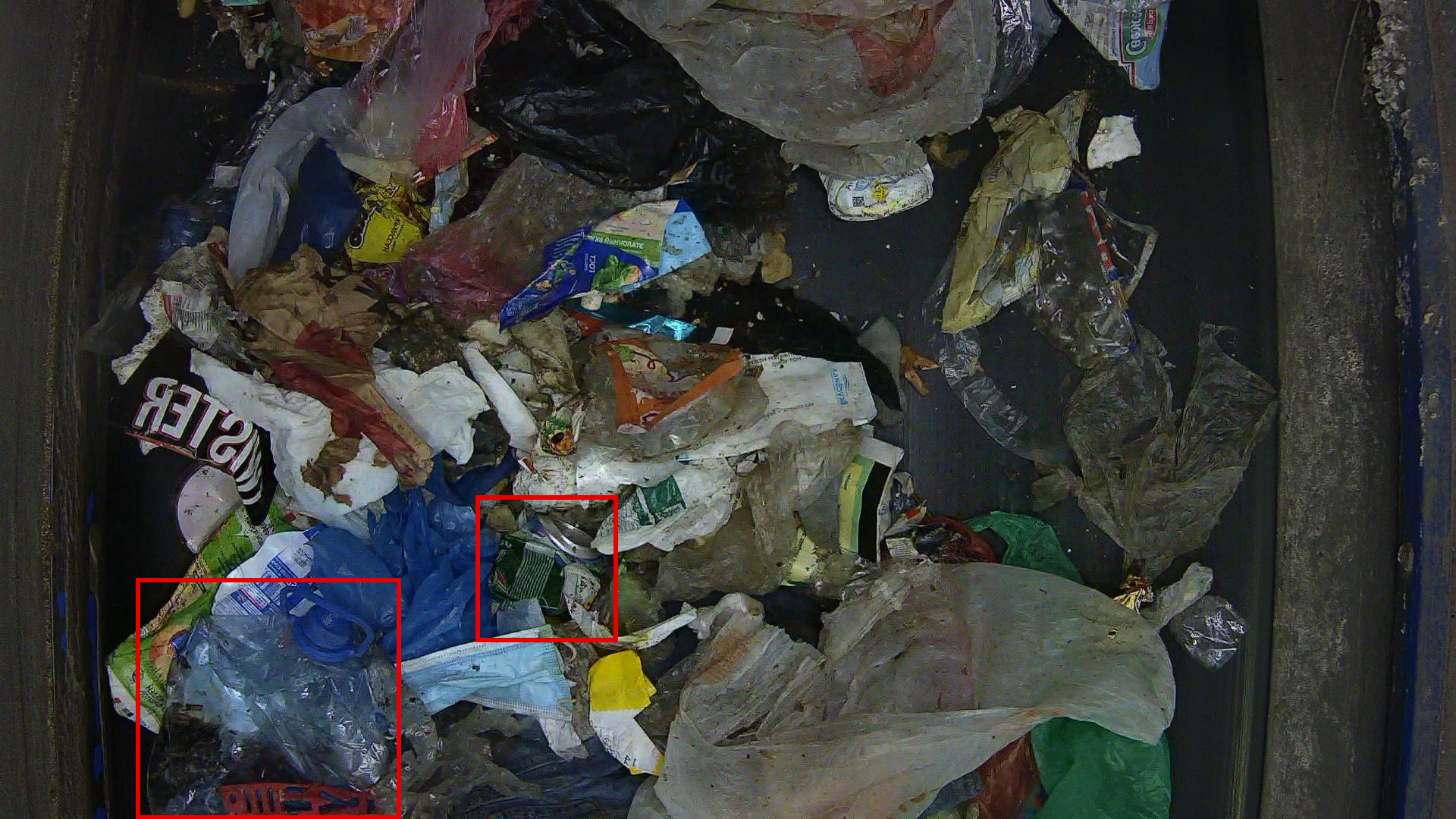}
\hfill
\includegraphics[width=0.48\textwidth]{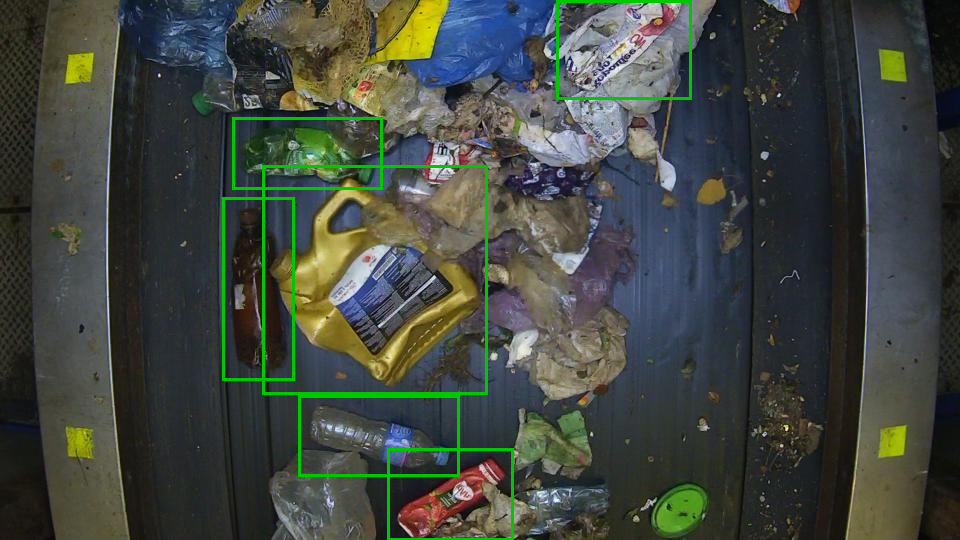}
\caption{WaRP-D, the one dataset where density and recall move together in the unexpected direction. Left: a low-annotated-count image (2 ground-truth boxes, red, both missed) where the annotated items are visually inseparable from a cluttered, unannotated background. Right: a high-annotated-count image (6 boxes, green, all detected) of distinct, well-separated items on a clean conveyor surface.}\label{fig:warp-density}
\end{figure}
\FloatBarrier
\end{document}